\documentclass[11pt]{article}

\usepackage[preprint]{acl}
\usepackage{fontawesome}
\usepackage{longtable}
\usepackage{amsmath}

\usepackage{times}
\usepackage{color}
\usepackage{latexsym}
\usepackage{CJKutf8}
\usepackage[T1]{fontenc}

\usepackage[utf8]{inputenc}

\usepackage{microtype}
\usepackage{multirow}
\usepackage{booktabs}
\usepackage{colortbl}
\usepackage{algorithm}
\usepackage{algorithmic}
\usepackage{inconsolata}

\usepackage{graphicx}

\title{AutoPatent: A Multi-Agent Framework for Automatic Patent Generation}

\author{
\textbf{Qiyao Wang\textsuperscript{1,2}}\thanks{Equal Contribution},
  \textbf{Shiwen Ni\textsuperscript{1}}\footnotemark[1],
   \textbf{Huaren Liu\textsuperscript{2}},
       \textbf{Shule Lu\textsuperscript{2}},
    \textbf{Guhong Chen\textsuperscript{1,3}},\\
      \textbf{Xi Feng\textsuperscript{1}},  \textbf{Chi Wei\textsuperscript{1}},
      \textbf{Qiang Qu\textsuperscript{1}}, \textbf{Hamid Alinejad-Rokny\textsuperscript{5}},
  \textbf{Yuan Lin\textsuperscript{2}}\thanks{Min Yang and Yuan Lin are corresponding authors.},
  \textbf{Min Yang\textsuperscript{1,4}}\footnotemark[2]
\\
  \textsuperscript{1}Shenzhen Key Laboratory for High Performance Data Mining,\\
  Shenzhen Institute of Advanced Technology, Chinese Academy of Sciences\\
  \textsuperscript{2}Dalian University of Technology,
  \textsuperscript{3}Southern University of Science and Technology\\
  \textsuperscript{4}Shenzhen University of Advanced Technology, \textsuperscript{5}The University of New South Wales\\
      \small{wangqiyao@mail.dlut.edu.cn, zhlin@dlut.edu.cn}, \small{\{sw.ni, min.yang\}@siat.ac.cn}}

\begin{document}
\maketitle
\begin{abstract}

As the capabilities of Large Language Models (LLMs) continue to advance, the field of patent processing has garnered increased attention within the natural language processing community. However, the majority of research has been concentrated on classification tasks, such as patent categorization and examination, or on short text generation tasks like patent summarization and patent quizzes. In this paper, we introduce a novel and practical task known as \textbf{Draft2Patent}, along with its corresponding D2P benchmark, which challenges LLMs to generate full-length patents averaging 17K tokens based on initial drafts. Patents present a significant challenge to LLMs due to their specialized nature, standardized terminology, and extensive length. We propose a multi-agent framework called \textbf{AutoPatent} which leverages the LLM-based planner agent, writer agents, and examiner agent with PGTree and RRAG to generate lengthy, intricate, and high-quality complete patent documents. The experimental results demonstrate that our AutoPatent framework significantly enhances the ability to generate comprehensive patents across various LLMs. Furthermore, we have discovered that patents generated solely with the AutoPatent framework based on the Qwen2.5-7B model outperform those produced by larger and more powerful LLMs, such as GPT-4o, Qwen2.5-72B, and LLAMA3.1-70B, in both objective metrics and human evaluations. We will make the data and code available upon acceptance\footnote{https://github.com/QiYao-Wang/AutoPatent}.
\end{abstract}

\section{Introduction}

\begin{figure}[t]
  \includegraphics[width=1\linewidth]{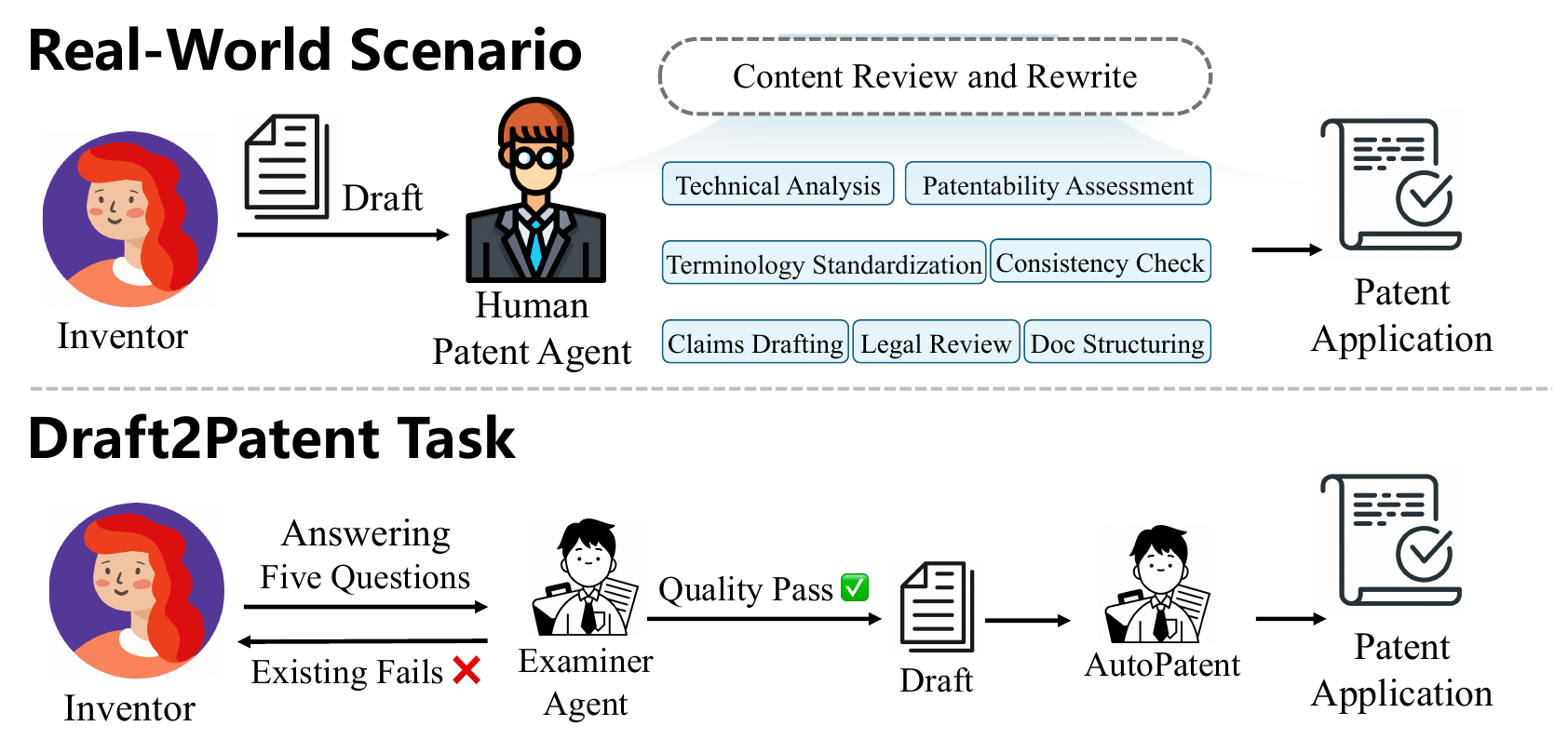}
  \caption {Draft2Patent Task. Automating patent drafting by simulating real-world scenarios.}
   \label{figure1}
\end{figure}

As a representative of Intellectual Property (IP), a patent is an exclusive right granted for an invention, which can benefit inventors by providing them with legal protection of their inventions\footnote{https://www.wipo.int/web/patents}. The inventor should draft a patent and submit it to a national or regional intellectual property (IP) office, such as the United States Patent and Trademark Office (USPTO) or the European Patent Office (EPO), to obtain a patent grant \citep{toole2020inventing}. The patent will be examined by a patent examiner for patentability. The examiner decides whether the proposed invention is useful, non-obvious, and statutory, and searches for prior arts within the technology field of the invention to confirm whether it is novel \citep{41638, 17847}. Therefore, inventors should draft a detailed, embodied patent and maximize the legal scope of protection for the invention without infringing on other patents. 

A patent typically consists of a title, abstract, background, summary, detailed description and claims \citep{wipo2022}. The drafting of patent is usually carried out by a human patent agent who is familiar with patent law and has passed the patent bar exam. Patent covers various technical fields, which requires human patent agents to possess a broad knowledge base. However, this process is still entirely conducted manually which results in high labor and time costs and lowering efficiency.

We introduce a novel real-world task named \textbf{Draft2Patent} for converting an inventor's draft into a complete patent, as shown in Figure~\ref{figure1}. We construct a challenging benchmark  named \textbf{D2P} for this task, which contains 1,933 draft-patent pairs and other patent metadata. The source patents are derived from the HUPD dataset \citep{NEURIPS2023_b4b02a09}, and we generate drafts meeting the quality requirements by interacting with GPT-4o-mini \citep{achiam2023gpt} using five specific questions. There are mainly two challenges: the average length of drafts and patents in D2P are exceed 4K and 17K tokens, and a patent must ensure its content is both patentable and compliant with technical and legal standards. 

With the development of Large Language Models (LLMs), LLM-based agents have demonstrated their powerful capabilities in understanding, planning, memory, rethinking, and action \citep{Cheng2024ExploringLL} in many knowledge-intensive domains, such as biomedicine \citep{Gao2024, kim2024mdagents}, finance \citep{Yu2023FinMemAP,10.1145/3637528.3671801,Yang2024FinRobotAO}, education \citep{ni2024educational} and law \citep{Cui2023ChatlawAM, Sun2024LawLuoAC, chen2024agentcourt, sun2024lawluo}. LLM-based agents can meet the knowledge demands of patent drafting in both law and technical fields. We propose a multi-agent framework for automatic patent drafting named \textbf{{AutoPatent}}, which can generate a complete patent using specialized expert agents. 

We experiment on the D2P benchmark using commercial models such as GPT-4o, GPT-4o-mini \citep{achiam2023gpt} and open source models such as LLAMA3.1 series (8B and 70B) \citep{dubey2024llama}, Qwen2.5 series (7B, 14B, 32B and 72B) \citep{qwen2}, and Mistral-7B \citep{jiang2023mistral}. Compared to other approaches, Qwen2.5-7B with AutoPatent demonstrates outstanding performance, achieving higher scores on objective metrics and human evaluation while significantly reducing repetition errors. Our main contributions include:
\begin{itemize}
	\item We introduce a new task with high application value, Draft2Patent, and construct the corresponding D2P benchmark, which contains 1,933 draft-patent pairs and requires the LLM to generate complete patent documents with an average length of 17K tokens.
	\item We present AutoPatent, an innovative multi-agent framework that leverages the collaborative efforts of LLM-based planning agents, writing agents, and examining agents for automatically producing high-quality patents.
	\item We propose two innovative methods, \textbf{PGTree} (\textbf{P}atent Writing \textbf{G}uideline \textbf{Tree}) and \textbf{RRAG} (\textbf{R}eference-\textbf{R}eview-\textbf{A}ugmented \textbf{G}eneration), and ablation experiments demonstrate the effectiveness of these two modules.
	\item  Numerous experiments prove that our AutoPatent excellent in both objective metrics and human evaluation. Moreover, our AutoPatent framework has nice migration and generalization properties, which can significantly improve the patent generation capability of various LLMs.

\end{itemize}

\section{Related Work}

\paragraph{Patent Writing.} Researchers have processed the text structure in patent domain with multiple nature language processing methods. \citep{ni-etal-2024-mozip, wang2024ipeval} focus on the application of patent law-related question-answering task in real-world scenario of intellectual property field. \citep{jiang2024artificial} summarizes the patent-related tasks into two types: patent analysis and patent generation. The patent generation task typically includes summarization \citep{10.1007/978-3-030-30244-3_42, sharma-etal-2019-bigpatent}, translation \citep{wirth2023building, heafield-etal-2022-europat}, simplification \citep{Casola2023CreatingAS}, and patent writing.

	The patent writing task previously focused on the internal conversion of a patent. \citep{LEE2020101983} preliminarily validated the feasibility of using GPT-2-based \citep{radford2019language} language models to construct patent claims. \citep{10.1145/3340531.3418503} converted patent abstract to claims through fine-tuning transformer-based models. \citep{jiang2024can} introduce a task for generating claims based on detailed description and constructed a benchmark to test this capability of LLMs. \citep{zuo-etal-2024-patenteval} use LLMs to convert claims into abstract and generate subsequent independent or dependent claims from existing claims. 
	
	But these tasks did not focus on writing a complete patent, \citep{knappich2024pap2pat} constructed a dataset with paper-patent pairs based on chunk-based, outline-guided method to convert papers into patents. But in real-world scenarios, patent granting is affected by previously published papers. Our Draft2Patent task focuses more on interaction between inventors and patent agents, aiming to generate a complete and high-quality patent that can even be submitted to the IP office.

\paragraph{LLMs-based Multi-Agent Framework for Long Text Generation.}\citep{NEURIPS2023_b4b02a09} revealed that the average length of patents' detailed description exceeds 10k tokens, making it challenging to generate. To solve the difficult and valuable task, researchers constructed many multi-agent framework based on role-playing or customized collaboration process \cite{ijcai2024p0890,hong2024metagpt, Chen2024FromPT}. \citep{bai2024longwriter} indicated that the limitations on LLMs’ output length stem from the long-tail distribution of the training dataset’s length. And they proposed a writing pipeline to solve it based on the agents' capabilities of planning. \citep{shao-etal-2024-assisting} focused on the generation of wikipedia-like articles based on the agents' brainstorm before writing. And \citep{huot2024agents} divided the story writing task into the specific agent's writing task and designed a evaluation framework to asses long narratives.

\begin{figure*}[t]
\centering
  \includegraphics[width=1\linewidth]{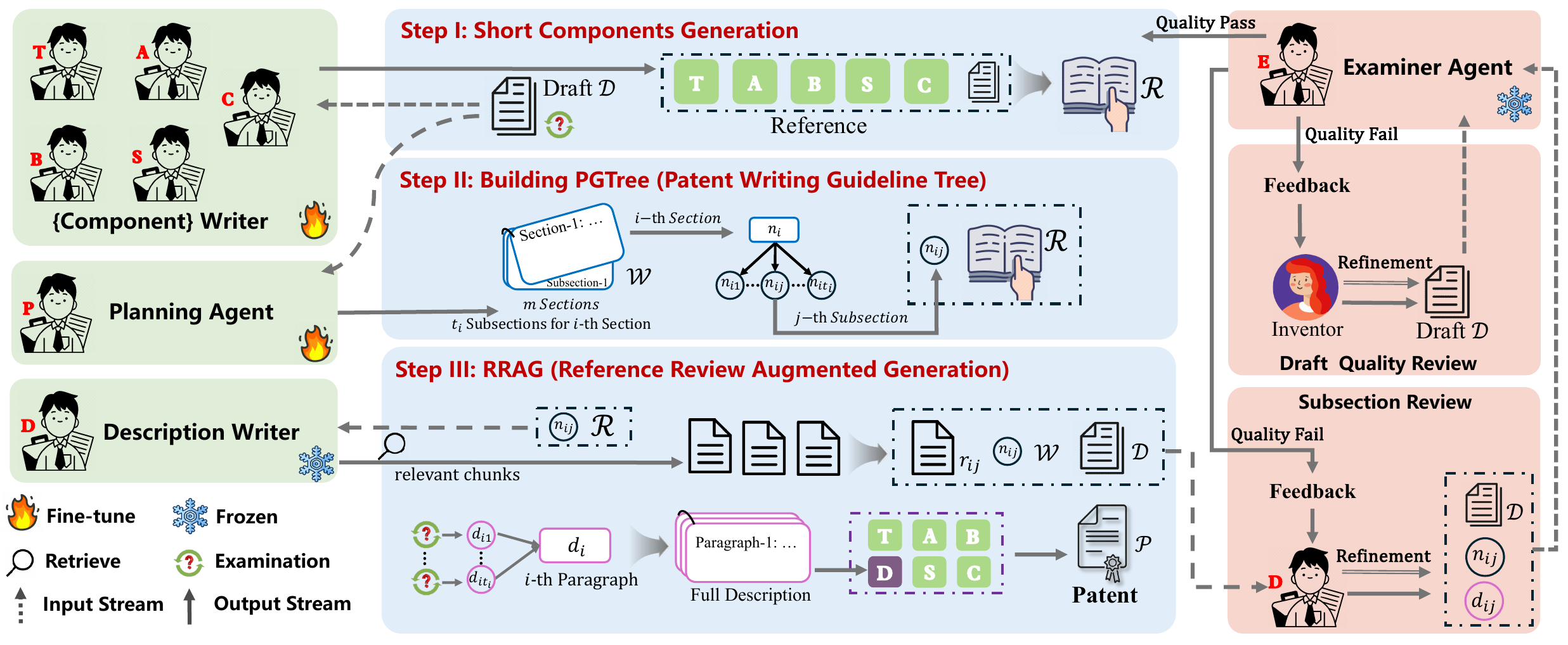}
  \caption {An overview of the AutoPatent framework, which includes eight agents and four steps for automatically generating a patent $\mathcal{P}$ from a draft $\mathcal{D}$.  The fire represents the corresponding agents' parameters are fine-tuned, while the snowflake represents the agents' parameters are frozen. The magnifying glass represents using the description writer for retrieval, while the question mark indicates using the examiner agent for text review. The dashed arrows represent the model’s input stream, while the solid arrows represent the model’s output stream.}
   \label{figure2}
\end{figure*}

\section{Draft2Patent Task}

The IP office only publishes granted patents of inventors, while inventors and patent agents never make their drafts public. In this section, we introduce our novel patent drafting task Draft2Patent and the agent-based method we used to construct the D2P benchmark, in detail.

\subsection{Task Definition}

We define Draft2Patent task by simulating real-world scenarios, as shown in Figure~\ref{figure1}. In real scenario, inventors usually deliver a patent technical draft, which contains the most comprehensive information, to a human patent agent, and ask them to draft a high-quality patent. The patent agent will review and rewrite the patent in terms of patentability, terminology standardization, terminology consistency, claim drafting and legal compliance. This process typically costs a lot of time before the patent is submitted to the IP office.

We leverage an agent-based method to simulate the interaction between inventors and patent agents, designing five questions  $q_1, q_2, \dots, q_5$ that encompass all the relevant information about an invention. We combine them with the inventors' answers $a_1,a_2,...,a_5$ to form the patent draft $\mathcal{D}$ and then we can use it to generate the patent $\mathcal{P}$:
	\begin{equation}
		\mathcal{D}=\{(q_1,a_1),(q_2,a_2),...,(q_5,a_5)\}
	\end{equation}
	Where five questions $q_1,q_2,...,q_5$ are as shown in Appendix~\ref{A1:appendix}. And where the $\mathcal{P}$ consists of the title, abstract, background, summary, claims, and detailed description of a patent.

\subsection{D2P Dataset Construction}
\label{Section3.2}

\paragraph{Draft Construction.}We use the HUPD \citep{NEURIPS2023_b4b02a09} dataset constructed based on USPTO’s publicly available patent as data source. We randomly select granted patent samples labeled with \textit{decision} as \textit{ACCEPTED}, ensuring that the contents of them are complete. For the patent $\mathcal{P}$, we simulate GPT-4o-mini as the inventor, asking it the five questions $q_1,q_2,...,q_5$.
The corresponding answers $a_1,a_2,...,a_5$ are then combined to form the draft.

\paragraph{Draft Quality Review.}After obtaining 2,000 patent drafts, we establish a standard for assessing the quality of the answers to each question $q_i$ to ensure they contain sufficient information about the invention. We simulate GPT-4o as a patent examiner agent to evaluate whether the answers fully address these questions, the concrete prompt as shown in Appendix~\ref{A2:appendix}. Finally, we obtain 1,933 patent drafts that meet the quality standard through the collaborative assessment of the LLM patent agent and human patent agent. These drafts are then divided into a training set of 1,500, a validation set of 133, and a test set of 300.

\paragraph{Other Metadata Construction.} Our D2P dataset not only contains draft-patent pairs, but also includes a fine-tuning dataset for short components generation and patent writing guideline tree (PGTree) generation. We combine the metadata in HUPD with drafts to create paired data, such as draft-title pairs. For patent $\mathcal{P}$, we simulate GPT-4o-mini as a assistant to summarize each part of the $\mathcal{P}$’s description, using it as a PGTree $\mathcal{W}$ for writing the detailed description. 

\begin{table}[b]
\centering
\caption{Text length distribution in the D2P benchmark.}
\label{table1}
\scalebox{0.6}{\begin{tabular}{lrrrr} 
\hline
\textbf{Section} & \textbf{All \#Tokens} & \textbf{Train \#Tokens} & \textbf{Valid \#Tokens} & \textbf{Test \#Tokens}  \\ 
\hline
Title            & 16.2                           & 16.1                        & 16.3                        & 16.2                        \\
Abstract         & 158.7                          & 159.8                       & 151.9                       & 156.7                       \\
Claims           & 1295.1                         & 1292.9                      & 1279.5                      & 1313.1                      \\
Summary          & 1287.5                         & 1300.7                      & 1103.9                      & 1303.1                      \\
Background       & 598.2                          & 595.4                       & 589.9                       & 616.1                       \\
Description & 14081.4                        & 14210.9                     & 12725.9                     & 14035.0                     \\
PGTree             & 1356.8                         & 1352.6                      & 1373.6                      & 1344.3                      \\
Draft            & 4076.5                         & 4078.0                      & 4022.5                      & 4082.9                      \\
Patent           & 17005.3                        & 17619.9                     & 15911.6                     & 17484.3                     \\
\hline
\end{tabular}}
\end{table}

We calculate the average length of the 1,933 drafts, patents, PGTrees, and other metadata in D2P benchmark using GPT-4o-mini's tokenizer, as shown in Table~\ref{table1}. The average length of a complete patent exceeds 17K tokens, with the detailed description averaging over 14K tokens and accounting for more than 80\% of the total, while each remaining section averages less than 2K tokens.

\section{AutoPatent Framework}

We propose an automatic multi-agent patent drafting framework named \textbf{{AutoPatent}} for Draft2Patent, as shown in Figure~\ref{figure2}. We design a specialized pipeline with eight agents and three steps to simulate the process of patent drafting in real-world scenario. We use five specialized agents to generate various sections of a patent, assigning a dedicated writer agent to each short component. And we assign a planning agent to generate a two layers, multi-way PGTree that instruct the reference-review-augmented generation (RRAG) process. We also assign an examiner agent to evaluate the quality of the generated subsections and provide modification suggestions.

In this section, we introduce each agent and the workflow of our AutoPatent framework for drafting the patent $\mathcal{P}$ in detail, given the draft $\mathcal{D}$. We also provide Algorithm~\ref{alg:autopatent}, which outlines the key steps of AutoPatent framework.
\begin{algorithm}[t]
\small
\caption{AutoPatent Framework.}
\label{alg:autopatent}
\renewcommand{\algorithmicrequire}{\textbf{Input:}}
	\renewcommand{\algorithmicensure}{\textbf{Output:}}
\begin{algorithmic}[1]

\REQUIRE Draft $\mathcal{D}$
\ENSURE  Complete patent $\mathcal{P}$
\STATE T for Title, A for Abstract, B for Background, S for Summary and C for Claims;
\STATE T, A, B, S, C = componentWriter.write($\mathcal{D}$)
\STATE Reference $\mathcal{R}$ = \{T, A, B, S, C, $\mathcal{D}$\}
\STATE PGTree $\mathcal{W}$ = planningAgent.plan($\mathcal{D}$) with $m$ sections, while the $i$-th section has $t_i$ subsections;
\FOR{$i \gets 1$ to $m$}
\FOR{$j \gets 1$ to $t_i$}
\STATE Retrieval content $r_{ij}$ from $\mathcal{R}$ with guideline $n_{ij}$;
\STATE $r_{ij}$ = descriptionWriter.retrieve($n_{ij}$, $\mathcal{R}$)
\STATE Description writer write the subsection $d_{ij}$;
\STATE $d_{ij}$ = descriptionWriter.write($r_{ij}$, $n_{ij}$, $\mathcal{W}$, $\mathcal{D}$)
\STATE Examiner Agent review $d_{ij}$ and give feedback;
\STATE Review, Feedback = examinerAgent.review($d_{ij}$)
\FOR{Review is Fail}
\STATE $d_{ij}$ = descriptionWriter.refine($d_{ij}$, Feedback)
\STATE Review, Feedback=examinerAgent.review($d_{ij}$)
\ENDFOR
\ENDFOR
\ENDFOR
\STATE Detailed Description D = $\{d_{ij}|1\le j \le t_i, 1\le i\le m\}$
\STATE $\mathcal{P}$ = concat(T, A, B, S, D, C)
\RETURN Complete patent $\mathcal{P}$
\end{algorithmic}
\end{algorithm}

\subsection{Agents}

We define each agent $\mathcal{A}$ as a sequence-to-sequence model that takes text as input and generates text as output. In AutoPatent framework, we design eight agents, categorized into three types: writer, planner, and examiner. Each agent has its own task and set of instructions.

\paragraph{Writer Agent.}We categorize six writer agents into two types: short component writer and description writer. Different parts of a patent exhibit significant stylistic differences, with the abstract typically being a single short paragraph and the claims often being lengthy and structured with numbered points. As shown in Table~\ref{table1}, all the average length of component is less than 2K tokens, except detailed description. 

We define these agents as $\mathcal{A}_i$, where $i\in \{T, A, B, S, C, D\}$, representing the corresponding component writers responsible for generating specific texts, such as the title writer. The description writer $\mathcal{A}_D$ is responsible for drafting each subsection of the description and executes the RRAG by retrieving references and interacting with the examiner agent to refine the content.

\paragraph{Planner Agent.}The average length of the detailed description exceeds 14,000 tokens, as shown in Table~\ref{table1}. It is currently impossible for LLMs to generate high-quality descriptions that meet legal and technical standards in a single pass. Leveraging the agent’s ability to organize and structure content, we define a planning agent $\mathcal{A}_P$ to generate PGTree for the detailed description, which serve to instruct the description writer in RRAG.

\paragraph{Examiner Agent.}We define a patent examiner agent, $\mathcal{A}_E$, for two types of quality assessments. When an inventor submits a new draft, the examiner agent needs to evaluate its quality to determine whether it meets the required standards and work collaboratively with the inventor to refine the draft, as detailed in Section~\ref{Section3.2}. When the description writer completes the generation of a subsection, the examiner agent actively intervenes to assess the quality of the content and provide feedback to the description writer for refinement until the review is successfully passed.

\begin{table*}[t]
\centering
\caption{Experiment results for objective metric. The \textbf{bold number} indicates the best performance under the same base model conditions, while the \underline{underlined number} represents the second-best. The \textcolor{red}{\textbf{red number}} represents the IRR scores below 90 when $t = 0.4$ and below 80 when $t = 0.2$.} 
\label{table2}
\scalebox{0.833}{\begin{tabular}{lccccccc} 
\hline
\textbf{Models}  & \textbf{BLEU}                              & \textbf{ROUGE-1}                      & \textbf{ROUGE-2}                      & \textbf{ROUGE-L}   &\textbf{IRR (t=0.2)} &\textbf{IRR (t=0.4)}                                       & \textbf{Avg \#Tokens}  \\ \hline
LLAMA3.1-8B& 10.83 &27.63 &10.96 &11.89 & 86.06 &96.45  &6431.17\\
LLAMA3.1-8B + SFT& 39.62 &30.82 &{15.65} &{19.44} &  \textcolor{red}{\textbf{49.17}} & \textcolor{red}{\textbf{64.90}}  &17052.23\\
LLAMA3.1-70B& 3.27 &28.80 &11.76 &10.95 & 88.41 &97.38  &1999.40\\
\hline
Qwen2.5-7B& 8.51 &30.61 &12.51 &12.77 & 89.49 &98.16 &2927.31\\
Qwen2.5-7B + SFT& 49.10 &\textbf{36.82} &\textbf{19.12} &\textbf{22.60} &   \textcolor{red}{\textbf{71.18}} & \textcolor{red}{\textbf{86.58}}  &11716.18\\
Qwen2.5-14B&5.70 &30.08 &12.01 &11.41 & 91.33 &98.48  &2480.07\\
Qwen2.5-32B& 2.65 &27.49 &10.95 &10.06 & 88.22 &98.06   &1916.09\\
Qwen2.5-72B& 15.10 &{33.09} &13.54 &13.86 & 91.14 &98.58 &3804.12\\
\hline
Mistral-7B& 2.46 &25.75 &9.66 &8.90 & 89.70 &98.04   &1703.09\\
Mistral-7B + SFT& 46.73  &\underline{33.18} &\underline{16.85} &\underline{21.14} &  \textcolor{red}{\textbf{62.49}} & \textcolor{red}{\textbf{81.11}}  &15031.85 \\
\hline
GPT-4o& 0.94 &23.26 &7.90 &6.96 & 90.60  &\underline{99.02} &1247.49\\
GPT-4o-mini& 3.32 &27.28 &10.06 &9.17 & 91.48 &98.94  &1935.26\\
\hline
\textit{\textbf{AutoPatent}}$_{\rm LLAMA3.1-8B}$& {49.22}  &{31.95} &{13.72} &{18.75} & {89.54} &96.07   &12496.74 \\
\textit{\textbf{AutoPatent}}$_{\rm Qwen2.5-7B}$& \textbf{53.03} &31.68  &{13.76}  &{19.08}  & \underline{93.61}  &98.79  &11432.79 \\
\textit{\textbf{AutoPatent}}$_{\rm Mistral-7B}$& 47.56  &30.40 &12.64 &17.72 &82.50  & 91.55  &15481.03 \\
\textit{\textbf{AutoPatent}}$_{\rm GPT-4o-mini}$& \underline{50.83}  &{30.85} &{11.24} &{16.37} & \textbf{96.92} &\textbf{99.56}  &13018.17\\
\hline
\end{tabular}}
\label{table2}
\end{table*}

\subsection{Framework Workflow}

As shown in the blue section of Figure~\ref{figure2}, we divide the workflow of the AutoPatent framework into three steps, simulating the real-world scenario of patent drafting.

\paragraph{Short Components Generation.}
\label{Section41}
In step \uppercase\expandafter{\romannumeral1}, we leverage different agents to generate various short components of a patent based on draft $\mathcal{D}$, considering differences in style. For open-source models with a parameter size of less than 14B, we fine-tune them using the D2P training set to enhance their ability to generate high-quality short components, while zero-shot prompting is used for commercial or larger models. The concrete prompt for supervised fine-tuning of short component generation is shown in Appendix~\ref{B1:appendix}. We then combine the generated title, abstract, background, summary, and claims with the draft $\mathcal{D}$ to form the reference $\mathcal{R}$, which can offer useful information for detailed description generation.

\begin{figure}[t]
  \includegraphics[width=1\linewidth]{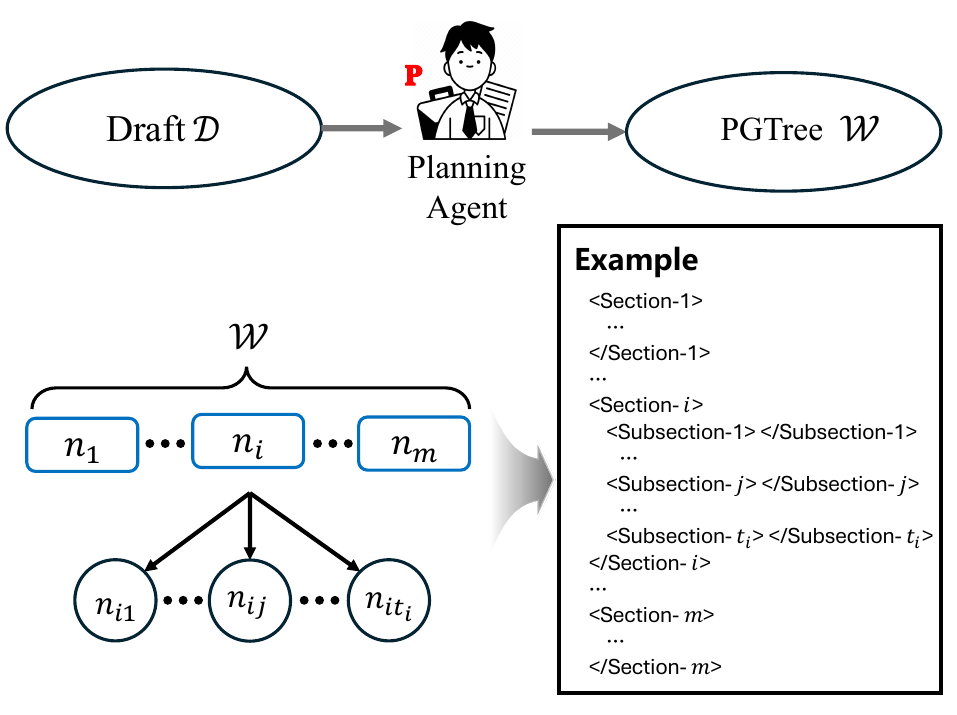}
  \caption {PGTree structure. The input of the planning agent is the draft $\mathcal{D}$, while the output is a PGTree $\mathcal{W}$, represented as a two-layer multiway tree.}
   \label{figure3}
\end{figure}

\paragraph{Patent Writing Guideline Tree (PGTree) Building.}In step \uppercase\expandafter{\romannumeral2}, given draft $\mathcal{D}$, we fine-tune the planning agent for smaller models and use zero-shot prompting for others, to generate a PGTree $\mathcal{W}$ for a detailed description. Assuming that the a PGTree consist of $m$ sections, the $i$-th section contains $t_i$ subsections, as shown in Figure~\ref{figure3}. The generated PGTree is structured as a two-layer multi-way tree, dividing the description generation task into two levels of outlines. The first level of the outline provides an overview of the section, while the second level offers concrete instructions for the description writer to generate concise content for that part of the description. The prompt as shown in Appendix~\ref{B2:appendix}. 

\paragraph{Reference-Review-Augmented Generation (RRAG).}In step \uppercase\expandafter{\romannumeral3}, given the guideline of $n_{ij}$, the description writer retrieves useful information $r_{ij}$ from reference $\mathcal{R}$. This process connects the planning agent with other short component writers, enhancing consistency throughout the entire patent and reducing the difficulty of tasks like writing claims in the detailed description. The prompt for retrieval as shown in Appendix~\ref{B1:appendix}.

After obtaining $r_{ij}$, we combine it with $n_{ij}$ and the PGTree $\mathcal{W}$ to instruct the description writer in generating the corresponding part $d_{ij}$ of the description.  The examiner agent then actively intervenes to review the quality of $d_{ij}$ and provides feedback for its refinement through multi-turn interactions with the description writer until the examiner agent deems the output acceptable. After traversing the PGTree for the $m$ sections, all the accepted $d_{ij}$ are concatenated to form the complete detailed description $d$. Finally, we combine all the generated text to form a complete patent $\mathcal{P}$. The description generation prompt is provided in Appendix~\ref{B3:appendix}, while the review prompt is in Appendix~\ref{examiner:appendix}.

\section{Experiment}
\subsection{Evaluation Metric}

We use the n-gram-based metric, BLEU \citep{papineni-etal-2002-bleu}, the F1 scores of ROUGE-1, ROUGE-2, and ROUGE-L \citep{lin-2004-rouge} as the objective metrics. \citep{10.1145/3485766} indicated that n-gram-based metrics exhibit a preference for repeated n-grams and short sentences. We propose a new metric, termed {IRR} (Inverse Repetition Rate), to measure the degree of sentence repetition within the patent $\mathcal{P}=\{s_i|1\le i\le n\}$, which consists of $n$ sentences. The {IRR} is defined as:
\begin{equation}
\textit{IRR} (\mathcal{P}, t) = \frac{C_n^2}{\sum_{i=1}^{n-1} \sum_{j=i+1}^{n} f(s_i, s_j) + \varepsilon}
\end{equation}
Where $\varepsilon$ is a small value added for smoothing to prevent division by zero, and $t$ is threshold for determining whether two sentences, $s_i$ and $s_j$, are considered repetitions based on their Jaccard similarity $J$, calculated after removing stop words. The function $f(s_i, s_j)$ is defined as:
\begin{equation}
	f(s_i, s_j) =
	\begin{cases}
	1, & \text{if } J(s_i, s_j) \geq t, \\
	0, & \text{if } J(s_i, s_j) < t.
	\end{cases}
\end{equation}

We invite three experts who are familiar with the patent law and patent drafting to evaluate the quality of generated patent using a single-bind review. We provide them with the review standards in Appendix~\ref{B5:appendix} regarding accuracy, comprehensiveness, logic, clarity, coherence, and consistency. We shuffle the two patents so that the experts won’t know which one comes from our AutoPatent framework. Each expert is required to select the winner between the two patents or choose a tie, using a real patent as a reference. Due to the extensive length of patents, we opted not to use LLM-based evaluation methods as they often fail to provide biased and inaccurate results.

\begin{figure*}[htpb]
\centering
  \includegraphics[width=1\linewidth]{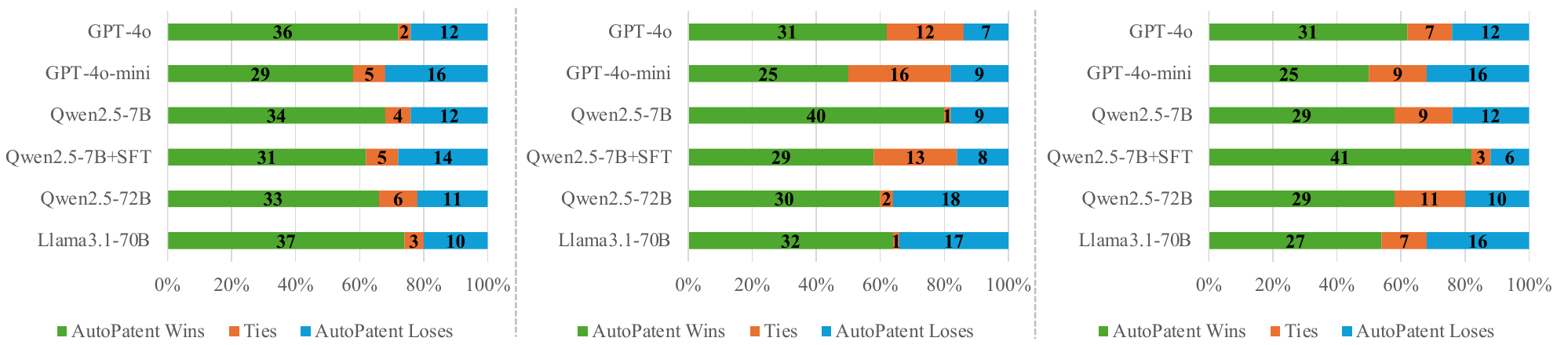}
  \caption {Human evaluation results. For each comparison, the left number indicates the count of AutoPatent wins, the middle shows the ties, and the right represents the cases where AutoPatent loses.}
   \label{figure4}
\end{figure*}

\subsection{Compared Method}
\label{Section5.2}

We compare our AutoPatent framework with two baseline methods: zero-shot prompting and supervised fine-tuning. Both of these methods take the draft as input and output a complete patent in an end-to-end manner.

\paragraph{Zero-shot Prompting Generation.}We use the zero-shot prompt shown in Appendix~\ref{B4:appendix} to instruct commercial models, including GPT-4o and GPT-4o-mini, and open-source models, including different sizes of LLAMA3.1, Qwen2.5 and Mistral to generate complete patents directly. We set the maximum token limit to all models, such as 16,384 for the GPT series models and 32,768 for the LLAMA3.1 series models. The temperature is set to 0.5 and top p to 0.9 for all models to improve the stability of the output. 

\paragraph{Supervised Fine-Tuning Generation.}We use 1,500 draft-patent pairs from D2P’s training set for fully supervised fine-tuning of the LLAMA3.1-8B, Qwen2.5-7B and Mistral-7B. We set the batch size to 2 and the learning rate to 2e-5 for 5 epochs, using 8 NVIDIA A800 80G GPUs and DeepSpeed ZeRO3, denoted as {+SFT} in table.

\subsection{Result}

\begin{table*}[!t]
\centering
\caption{Ablation results of ablation experiment results for removing PGTree or RRAG module.}
\label{table4}
\scalebox{0.9}{\begin{tabular}{llcccccc}
\hline
\textbf{Models} & \textbf{\textit{+AutoPatent}} & \textbf{BLEU} & \textbf{ROUGE-1} & \textbf{ROUGE-2} & \textbf{ROUGE-L} & \textbf{IRR (t=0.2)}&\textbf{Avg \#Tokens} \\
\hline
GPT-4o-mini &  & \textbf{50.83} & \textbf{30.85} & \textbf{11.24} & \textbf{16.37}  & 96.92 &13018.17 \\
 & \textit{w/o} PGTree & 3.43 & 27.61 & 10.27 & 9.23 & 91.05 &1914.74\\
 & {\textit{w/o} RRAG} & 48.39 & 30.76 & 11.21 & 15.94  &96.10  & 11426.72 \\
\hline
\end{tabular}}
\end{table*}

\paragraph{Objective Metric Results.}
We report the objective metric results on D2P's test set and the average length of generated patent in Table~\ref{table2}.
Observing the average length in the results, all models generate patents with an average length of less than 3,000 tokens using the zero-shot prompt. While generated with AutoPatent, the average length of the patent exceeds 10K tokens. For n-gram-based metric, our AutoPatent framework achieves the higher performance both within the same base models. We observed that when leveraging Qwen2.5-7B as the base model in the AutoPatent framework surpasses the performance of GPT-4o-mini.

The average IRR score across all real patents in the test set is 91.33 when t is 0.2 and 98.57 when t is 0.4. This phenomenon is primarily attributed to the stylistic characteristics of patent language, such as the inclusion of claims within the description. The patents generated using the supervised fine-tuning method exhibit significant repetition errors, with an IRR score of 49.17 for LLAMA3.1-8B{+SFT}, 71.18 for Qwen2.5-7B{+SFT} and 62.49 for Mistral-7B{+SFT} when the threshold is set to 0.2. These severe repetition errors lead to over-rewarding in n-gram-based metrics, resulting in the best scores for {SFT}, despite its actual quality being poor.

\paragraph{Human Evaluation Results.}

We report the human evaluation results in Figure~\ref{figure4}, comparing Qwen2.5-7B+AutoPatent with zero-shot prompting generation (denoted as GPT-4o, GPT-4o-mini, Qwen2.5-7B, Qwen2.5-72B, LLAMA3.1-70B) and SFT generation (denoted as Qwen2.5-7B+SFT) for 50 generated patents.
The three human experts all agree that the quality of the complete patents generated using the AutoPatent framework outperformed other models across six dimensions in Appendix~\ref{B5:appendix}.

\section{Analysis}	

\subsection{Ablation Study}
We conduct three types of ablation experiments. Two of these use GPT-4o-mini as the base model to evaluate the AutoPatent framework without PGTree or RRAG module. The third experiment uses LLAMA3.1-8B as the base model to evaluate the performance of the short components writer and planning agent without fine-tuning. The results are reported in Table~\ref{table4} and Table~\ref{table5}.

\paragraph{Ablation on PGTree.}We conduct an ablation experiment on PGTree. After the different short component writers generate the corresponding parts of the patent, the description writer completes the full detailed description in a single pass without utilizing the PGTree. Observing the results in Table~\ref{table4}, the average length drops below 2,000 tokens, and all objective metrics decrease significantly, with BLEU experiencing an almost 15-times reduction.

\paragraph{Ablation on RRAG.}We conduct an ablation experiment on RRAG. When the description writer generates a subsection $d_{ij}$ using the guideline $n_{ij}$, it simply adds it to the list of description candidates without considering advice from the examination agent. Table~\ref{table4} shows that removing the RRAG results in a 4.8\% reduction in the BLEU score, along with declines in all other objective metrics. Without the reference and the supervision of the examiner agent, repetition errors slightly increase, as reflected by a minor decrease in the IRR score.

\paragraph{Ablation on Fine-tuning Agent.}We conduct an ablation experiment on generating short components without fine-tuning. Due to differences among patents, the base model exhibits varying capabilities when generating different short components. As shown in Table~\ref{table5}, the objective metric scores demonstrate significant improvement, particularly in generating patent-style titles, with BLEU and ROUGE-L scores increasing by approximately 6.7 times. We also use BERTScore as a semantic metric, observing significant improvement across all tasks.

\begin{table}[htpb]
\centering
\caption{Ablation results on fine-tuning agents. +\text{SFT} denotes the fine-tuning agent, while the other row represents results without fine-tuning.}
\label{table5}
\scalebox{0.9}{\begin{tabular}{l c c c c}
\hline
\textbf{Task} & & \textbf{BLEU} & \textbf{ROUGE-L} & \textbf{BERTScore} \\
\hline
\multirow{2}{*}{D2T} & & 8.64          & 8.65         & 60.46        \\
                     & +SFT & \textbf{67.09} & \textbf{66.48} & \textbf{90.86} \\
\hline
\multirow{2}{*}{D2A} & & 38.75         & 23.16        & 63.47        \\
                     & +SFT & \textbf{62.52} & \textbf{38.66} & \textbf{72.05} \\
\hline
\multirow{2}{*}{D2B} & & 27.55         & 11.93        & 58.95        \\
                     & +SFT & \textbf{36.11} & \textbf{19.35} & \textbf{64.67} \\
\hline
\multirow{2}{*}{D2S} & & \textbf{29.49}         & 14.64        & 57.77        \\
                     & +SFT & 28.30 & \textbf{21.88} & \textbf{66.99} \\
\hline
\multirow{2}{*}{D2C} & & 39.41         & 24.24        & 70.69        \\
                     & +SFT & \textbf{48.72} & \textbf{30.49} & \textbf{74.09} \\
\hline
\multirow{2}{*}{D2W} & & 58.74         & 20.69        & 66.48        \\
                     & +SFT & \textbf{72.37} & \textbf{26.31} & \textbf{73.46} \\
\hline
\end{tabular}}
\end{table}

\subsection{Case study}

We carefully review all the generated patents based on different methods and conduct a detailed analysis. The generated patent using SFT exhibits significant repetition errors, resulting in meaningless content and even leading to the failure to generate complete content. This phenomenon results in high ROUGE scores for SFT, but human evaluation highlights its shortcomings. As shown in Appendix~\ref{D1}, we present a comparison between SFT and our AutoPatent framework.

The patents generated using AutoPatent exhibit greater comprehensiveness, with no missing parts, and are even capable of generating flowcharts. Other methods often fail to generate complete content, such as missing descriptions, as shown in Appendix~\ref{D2}. The AutoPatent framework also improves consistency in generated patents, such as ensuring alignment between claims in the description and those outside it, which other methods fail to maintain.

\section{Conclusion}

In this work, we introduce a novel and practical task, Draft2Patent and its corresponding D2P benchmark containing 1,933 draft-patent pairs, which requires LLMs to generate full patent documents with an average length of 17K tokens. Due to the specialized nature of patents, standardized terminology, and long length, mainstream LLMs perform poorly. We propose an innovative multi-agent framework, AutoPatent, which capitalizes on the collaborative efforts of a LLM-based planning agent, six writing agents, and a review agent to produce high-caliber patent documents with novel PGTree and RRAG method. 

Our experimental results indicate that AutoPatent markedly enhances the capacity of various LLMs to generate full patents. Moreover, we have discovered that patents generated exclusively with the AutoPatent framework, utilizing the Qwen2.5-7B model, surpass those produced by more extensive and potent LLMs, such as GPT-4o, Qwen2.5-72B, and LLAMA3.1-70B, in both objective metrics and human evaluations. Remarkably, the quality of patents generated by AutoPatent rivals that of human authorship. We hope that AutoPatent will revolutionize the way patents are generated and managed, simplifying the process and lowering the barriers to innovation.
\section*{Limitations}

The patent evaluation task is highly challenging, involving intricate legal and technical standards that demand meticulous review by human experts. This results in low efficiency and high costs in human evaluation, and we will explore a concise method for automated patent evaluation in the future. Due to limitations in computing resources, we do not fully fine-tune LLMs with a parameter size of 14B or larger.

\section*{Ethics Statement}
All patent data used in this paper are obtained from publicly accessible sources. The purpose of the Draft2Patent task is to improve the efficiency of patent agents in drafting applications before submission to the IP office. We do not encourage people to use our method to generate fake or meaningless patents that would burden the IP office’s examination department. We acknowledge that the patents generated by our method are not yet sufficient to be submitted directly to the IP office. They still require modification by a patent agent to ensure they meet patent law and technical standards.

\section*{Acknowledgement}
This work was supported by GuangDong Basic and Applied Basic Research Foundation (2023A1515110718 and 2024A1515012003), China Postdoctoral Science Foundation (2024M753398), Postdoctoral Fellowship Program of CPSF (GZC20232873).

\bibliography{custom}

\appendix

\section{Prompt for Data Collection}
\label{A:appendix}
\subsection{Five Questions}
\label{A1:appendix}

As shown in Table~\ref{5questions}, we present the five questions for draft collection. These questions arise from our discussions with professional human patent agents and contain comprehensive information for patent drafting.

\begin{table}[htpb]
\centering
\caption{Five key questions for obtaining a patent technical draft.}
\label{5questions}
\begin{tabular}{|p{7.5cm}|}
\hline\\
\parbox{1\linewidth}{
{\large\textbf{Question 1 $q_1$:} }What is the technical problem that this patent aims to solve?\\
{\large\textbf{Question 2 $q_2$:} }What is the technical background of this invention, the most similar existing solutions, and its advantages over these solutions?\\
{\large\textbf{Question 3 $q_3$:} }What is the detailed technical solution of the invention?\\
{\large\textbf{Question 4 $q_4$:} }What are the key points of the invention, and which points are intended to be protected?\\
{\large\textbf{Question 5 $q_5$:} }What is the detailed description of each figure individually?\\
}
\\
\hline
\end{tabular}
\end{table}

\subsection{Draft Quality Review Prompt}
\label{A2:appendix}
As shown in Table~\ref{draft-quality-review-prompt}, we assess the draft’s quality with the LLM-based examiner agent using this prompt.

\begin{table*}[htpb]
\centering
\caption{Prompt used by LLM-based examiner agent for quality review of patent technical drafts. Where $q_i$ represents the $i$-th question and $a_i$ represents the answer to the $i$-th question.}
\label{draft-quality-review-prompt}
\begin{tabular}{|p{15.5cm}|}
\hline\\
\parbox{1\linewidth}{

{\large\textbf{For Question 1 $q_1$:} }\\
\# Draft: \{$a_1$\}

\# Requirements: The text of this draft section must include the technical problem solved by the invention. If it is included, just return <Result> Pass </Result>; if it is not included, return <Result> Fail </Result>, and provide a detailed explanation in <Reason> waiting for filling </Reason>.

Please tell me if this section of the draft meets the quality standards.\\}

\parbox{1\linewidth}{

{\large\textbf{For Question 2 $q_2$:} }\\
\# Draft: \{$a_2$\}

\# Requirements: The text of this draft section must include the background of the technology, the existing technical solutions, the shortcomings of the existing technology, and the advantages of the present invention. If it is included, just return <Result> Pass </Result>; if it is not included, return <Result> Fail </Result>, and provide a detailed explanation in <Reason> waiting for filling </Reason>.

Please tell me if this section of the draft meets the quality standards.\\}

\parbox{1\linewidth}{

{\large\textbf{For Question 3 $q_3$:} }\\
\# Draft: \{$a_3$\}

\# Requirements: The text of this draft section must include a detailed technical solution, which should describe the specific technical means for implementing the invention. If it is included, just return <Result> Pass </Result>; if it is not included, return <Result> Fail </Result>, and provide a detailed explanation in <Reason> waiting for filling </Reason>.

Please tell me if this section of the draft meets the quality standards.\\}

\parbox{1\linewidth}{

{\large\textbf{For Question 4 $q_4$:} }\\
\# Draft: \{$a_4$\}

\# Requirements: The text of this draft section must include the description of the drawings for the invention, where each figure must correspond to its respective drawing description one by one. If it is included, just return <Result> Pass </Result>; if it is not included, return <Result> Fail </Result>, and provide a detailed explanation in <Reason> waiting for filling </Reason>.

Please tell me if this section of the draft meets the quality standards.\\}

\parbox{1\linewidth}{

{\large\textbf{For Question 5 $q_5$:} }\\
\# Draft: \{$a_5$\}

\# Requirements: The text of this draft section must include a detailed technical solution, which should describe the specific technical means for implementing the invention. If it is included, just return <Result> Pass </Result>; if it is not included, return <Result> Fail </Result>, and provide a detailed explanation in <Reason> waiting for filling </Reason>

Please tell me if this section of the draft meets the quality standards.\\}
\\
\hline
\end{tabular}
\end{table*}

\subsection{PGTree Collection Prompt}
\label{A3:appendix}

As shown in Table~\ref{plan-data}, we use this prompt to obtain the PGTrees, which are then used to fine-tune the planning agent.

\begin{table*}[htpb]
\centering
\caption{Prompt for PGTree collection. Where  $d$ represents the detailed description.}
\label{plan-data}
\begin{tabular}{|p{15.5cm}|}
\hline\\
\parbox{1\linewidth}{
Description: \{$d$\}

Based on the provided patent description text, summarize the key parts and provide detailed guidance for drafting the content of each part. Please output in the following format, with each section described in a single paragraph:

<Section-1> Main content and drafting points for this section </Section-1>

<Section-2> Main content and drafting points for this section </Section-2>

...

<Section-n> Main content and drafting points for this section </Section-n>

Ensure that each section is specific and cohesive, covering the full content of the patent description. Please strictly adhere to the required format.\\}
\\
\hline
\end{tabular}
\end{table*}

\section{Prompt for AutoPatent Framework}

\subsection{Short Component Writer Prompt}
\label{B1:appendix}

As shown in Figure~\ref{figure2} and Section~\ref{Section41}, agents need to generate the corresponding short component, using the prompt in short components generation process, which as shown in Table~\ref{short-prompt}.

\begin{table*}[htpb]
\centering
\caption{Prompt for generating short components used in the autoPatent framework. Where $\mathcal{D}$ represents the patent technical draft.}
\label{short-prompt}
\begin{tabular}{|p{15.5cm}|}
\hline\\
\parbox{1\linewidth}{
\textbf{Title Writer Prompt}

Draft: \{$\mathcal{D}_i$\}

Based on the above patent draft, please generate a patent title that complies with legal and patent regulations, and follows the format below:

<Title>the title of patent</Title>
\\
}

\parbox{1\linewidth}{
\textbf{Abstract Writer Prompt}

Draft: \{$\mathcal{D}$\}

Based on the provided patent draft, please generate a patent abstract that complies with legal and patent regulations, following the format below:

<Abstract>the abstract of patent</Abstract>
\\
}

\parbox{1\linewidth}{
\textbf{Background Writer Prompt}

Draft: \{$\mathcal{D}$\}

Please generate the detailed background information for the patent based on the above patent draft. 

The background information should include the technical field of the patent, provide an objective introduction to the existing technology relevant to the invention, and point out any deficiencies or issues in the existing technology. Additionally, summarize the purpose or motivation of the invention without disclosing specific details. Please avoid negative comments about the existing technology or others' patents. The content should be clear and concise, avoiding unnecessary complexity.

Please output in the following format:

<Background>the background information of patent</Background>
\\
}

\parbox{1\linewidth}{
\textbf{Summary Writer Prompt}

Draft: \{$\mathcal{D}$\}

Please generate the summary for the patent based on the above patent draft. The summary should provide a detailed overview of the invention, including the technical field, the problems in the prior art that the invention addresses, and the key technical features of the invention. The summary should explain how the invention solves the identified problems without delving into specific implementation details. Ensure the summary is clear, concise, and focused on the invention's main objectives and advantages.

Please output in the following format:

<Summary>the summary of the patent</Summary>
\\
}

\parbox{1\linewidth}{
\textbf{Claims Writer Prompt}

Draft: \{$\mathcal{D}$\}

Based on the patent draft, please generate patent claims that comply with legal and patent regulations.

The claims should be written in clear language, avoiding ambiguity or vague descriptions.The independent claims should cover the core technical features of the invention and should not rely on other claims. The dependent claims should supplement or limit the independent claims, referencing the relevant independent claims.The claims must focus on a single invention, ensuring the unity of the invention, and must be consistent with the content of the draft.

Ensure that the described invention possesses novelty, inventive step, and industrial applicability.The claims should clearly define the scope of the invention's protection through specific technical features (such as components, steps, or systems).Each claim should end with a complete sentence, be numbered sequentially, and have an appropriate scope—neither too narrow nor too broad.

Please strictly adhere to these guidelines when generating the patent claims and following the format below:

<Claims>the claims of patent</Claims>
\\
}
\\
\hline
\end{tabular}
\end{table*}
	
\subsection{Planner Agent Prompt}
\label{B2:appendix}

Figure~\ref{figure3} illustrates the detailed structure of the PGTree. The concrete prompt used by the planning agent is shown in Table~\ref{tree-plan}.

\subsection{Description Writer Prompt}
\label{B3:appendix}
As shown in Figure~\ref{figure3}, the description writer is responsible for RRAG: executing the retrieval process, generating description subsections based on the PGTtrees, and refining the subsections based on feedback from the examiner agent.

\paragraph{Retrieval Reference Prompt.}As shown in Table~\ref{lube-prompt}, this is the prompt for finding useful content in reference $\mathcal{R}$ based on the guideline $n_{ij}$.

\paragraph{Description Generation Prompt.}As shown in Table~\ref{description-prompt}, this is the prompt for generating subsections of detailed descriptions based on PGTrees and useful content retrieved from references.

\paragraph{Refinement Content Prompt.}As shown in Table~\ref{refine-prompt}, this is the prompt for the description writer to refine the subsection based on feedback from the examiner agent.

\subsection{Examiner Agent Prompt}
\label{examiner:appendix}

As shown in Figure~\ref{figure2}, when the description writer completes the generation of a subsection, the examiner agent actively intervenes to assess the quality of the content and provide feedback. The concrete prompt used by the examiner agent is shown in Table~\ref{review-prompt}.

\section{More Prompts}

\subsection{Zero-Shot Prompt}
\label{B4:appendix}

We conduct experiments on GPT-4o, GPT-4o-mini, LLAMA3.1-8B, and LLAMA3.1-70B using a zero-shot prompt, as detailed in Table~\ref{zero-shot-prompt}.

\subsection{Human Expert Evaluation Standard}
\label{B5:appendix}

Since the experts we invited are all familiar with patent law and patent drafting, we only provide them with the dimensions and corresponding definitions they should focus on during the evaluation process, without providing additional knowledge. Meanwhile, we inform the experts to disregard any external factors, carefully read the patent texts to be evaluated, and finally choose between options document 1, document 2, or indicate a tie. Below are the dimensions the experts need to focus on:

\begin{itemize}
	\item \textbf{Accuracy:} The patent text must ensure that every technical detail is correct, avoiding vague expressions. Parameters, structures, and processes should be described specifically and clearly to ensure that the invention can be technically realized. The terms used should align with the standards of the technical field to avoid ambiguity and ensure consistent and accurate expression. When describing technical features, overly narrow or restrictive language should be avoided to ensure that the scope of patent protection is not unnecessarily limited.
	\item \textbf{Logic:} The patent text must be structured in accordance with patent law requirements, ensuring that the application includes the necessary sections, such as abstract, background, summary, detailed description, and claims. The logical progression of the text should be natural, enabling the reader to gradually understand the background, the innovation, and the specific application of the invention. Each section should be logically connected to the preceding and following technical descriptions to ensure a clear and coherent presentation of the invention.
	\item \textbf{Comprehensiveness: }The patent text should comprehensively disclose the invention so that a person skilled in the art can understand and implement it, avoiding any omissions or vague descriptions. In addition to detailing specific embodiments, the patent text should use broad terms that cover various modifications and alternatives to maximize the scope of legal protection. Given that patent texts are generally lengthy, combining full disclosure with broad protection ensures both the implementability of the invention and its legal strength.
	\item \textbf{Clarity: }The patent text should strike a good balance between technical and legal aspects to ensure that the description of the technical solution is both clear and easy to understand. While the text needs to maintain a certain level of technical rigor and legal accuracy, the language should still be concise and clear, avoiding overly complex or lengthy sentence structures. Redundant descriptions should be avoided unless necessary to convey critical technical details. This clarity ensures that the patent examiner can quickly grasp the core content of the invention.
	\item \textbf{Coherent: }The patent text must express the invention precisely to avoid ambiguity or the use of vague terms. Each technical feature should be clearly understood, avoiding terms like “almost” or “approximately” that create uncertainty. The sections and paragraphs should be organized logically, ensuring smooth progression of ideas. Each section should be clearly distinguished to allow the examiner to progressively understand the overall invention.
	\item \textbf{Consistency: }The patent text must maintain consistency with the provided Real Patent, ensuring that the description of the technical solution is accurate and coherent. References to technical features should complement rather than contradict each other. Consistency across sections and in terminology should be preserved, ensuring a clear and coherent expression of the invention’s core technology. This consistency reduces the risk of invalidation and enhances the patent’s legal stability.
\end{itemize}

\begin{table*}[htpb]
\centering
\caption{Prompt for PGTrees generation by the Planner Agent in the AutoPatent Framework. Where $\mathcal{D}$ represents the patent technical draft.}
\label{tree-plan}
\begin{tabular}{|p{15.5cm}|}
\hline\\
\parbox{1\linewidth}{
Draft: \{$\mathcal{D}$\}

Based on the provided patent draft, I need you to help me write a detailed writing guide for the patent description.

This guide should consist of multiple sections, with each section providing guidance for writing a part of the patent description and including key points to cover for that section.

Please output in the following format:

<Section-1> Main content and key points for writing this section </Section-1>  

<Section-2> Main content and key points for writing this section </Section-2>

...

<Section-n> Main content and key points for writing this section </Section-n>

Ensure that each part of the guide is clear, specific and cohesive, covering the entire content of the patent description. Please strictly adhere to the required format.
\\}
\\
\hline
\end{tabular}
\end{table*}

\begin{table*}[htpb]
\centering
\caption{Prompt for reference retrieval process. Where $\mathcal{R}$ is the reference content, and $n_{ij}$ is the writing guideline.}
\label{lube-prompt}
\begin{tabular}{|p{15.5cm}|}
\hline\\
\parbox{1\linewidth}{

Reference Conetent: \{$\mathcal{R}$\}

Writing Plan: \{$n_{ij}$\}

According to the patent text writing plan, determine which of the following contents are necessary for writing this section, and copy the all relevant content without modifying or adding anything. 

Just output the needed information for drafting this subsection.\\}
\\
\hline
\end{tabular}
\end{table*}

\begin{table*}[htpb]
\centering
\caption{Prompt for generating subsections of detailed description based on PGTree and retrieved reference content. Where $r_{ij}$ is the reference content, and $n_{ij}$ is the subsection of writing plan $\mathcal{W}$.}
\label{description-prompt}
\begin{tabular}{|p{15.5cm}|}
\hline\\
\parbox{1\linewidth}{

<Reference>\{$r_{ij}$\}</Reference>

Writing Guideline Overview: \{$\mathcal{W}$\}

Subsection Writing Guideline: \{$n_{ij}$\}

Based on the content in <Reference></Reference> and the subsection writing guideline, please draft this subsection, ensuring that the description complies with legal and patent regulations. 

Just output this subsection of patent description, and don't output other content.
\\}
\\
\hline
\end{tabular}
\end{table*}

\begin{table*}[htpb]
\centering
\caption{Prompt for refining the subsection already written $d_{ij}$ based on the examiner agent’s feedback, where $n_{ij}$ is the subsection of the writing plan $\mathcal{W}$.}
\label{refine-prompt}
\begin{tabular}{|p{15.5cm}|}
\hline\\
\parbox{1\linewidth}{
Writing Guideline Overview: \{$\mathcal{W}$\}

Subsection Writing Guideline: \{$n_{ij}$\}

The subsection already written: \{$d_{ij}$\}

Feedback from Patent Examiner: \{Feedback\}

Based on the subsection writing guideline and the feedback, revise the subsection to ensure it complies with legal and patent regulations while addressing the examiner's concerns. Do not say anything else. Only output the revised subsection.
\\}
\\
\hline
\end{tabular}
\end{table*}

\begin{table*}[htpb]
\centering
\caption{The prompt for reviewing the subsection already written  $d_{ij}$ using examiner agent, where $\mathcal{D}$ is draft and $n_{ij}$ is the subsection of the writing plan $\mathcal{W}$.}
\label{review-prompt}
\begin{tabular}{|p{15.5cm}|}
\hline\\
\parbox{1\linewidth}{
Draft: \{$\mathcal{D}$\}

<WritingGuideline> \{$n_{ij}$\} </WritingGuideline>

<Content> \{$d_{ij}$\} </Content>

<Requirement> 

Accuracy should ensure that technical details are clear and precise, aligning with law and technical standards. Logic should follow a natural progression with a clear structure. Comprehensiveness should fully disclose all necessary information required by the writing guideline. Clarity should feature concise and easily understandable language, balancing technical and legal descriptions. Coherence should ensure smooth expression, avoiding any ambiguity or uncertainty. Consistency should maintain uniform terminology, align fully with the draft, and avoid any contradictions.

</Requirement>

Refer to draft and evaluate whether the content meets the requirement provided, based on the given writing guideline.

If it complies with the requirement and writing guideline, return <Result>Pass</Result>; if it does not comply, return <Result>Fail</Result>. 

And you must provide helpful and detailed advice in <Advice>waiting for filling</Advice> regardless of whether the result is Pass or Fail.\\}
\\
\hline
\end{tabular}
\end{table*}

\begin{table*}[htpb]
\centering
\caption{Prompt for Zero-Shot Experiment. Where $\mathcal{D}$ represents the patent technical draft.}
\label{zero-shot-prompt}
\begin{tabular}{|p{15.5cm}|}
\hline\\
\parbox{1\linewidth}{

Draft: \{$\mathcal{D}$\}

Format requirements:

<Patent>

<Title> the title of patent </Title>

<Abstract> the abstract of patent </Abstract>

<Background> the background of patent </Background>

<Summary> the summary of the patent </summary>

<Claims> the claims of patent </Claims>

<Full Description> the full description of patent </Full Description>

</Patent>

Please write a complete patent document based on the above patent draft, following the format requirements. The document should be professional, coherent, clear, and precise.
\\}
\\
\hline
\end{tabular}
\end{table*}

\section{Patent Case}
\label{D}

\subsection{Repetition Error Case}
\label{D1}
\begin{figure*}[htpb]
\centering
  \fbox{\includegraphics[width=1\linewidth]{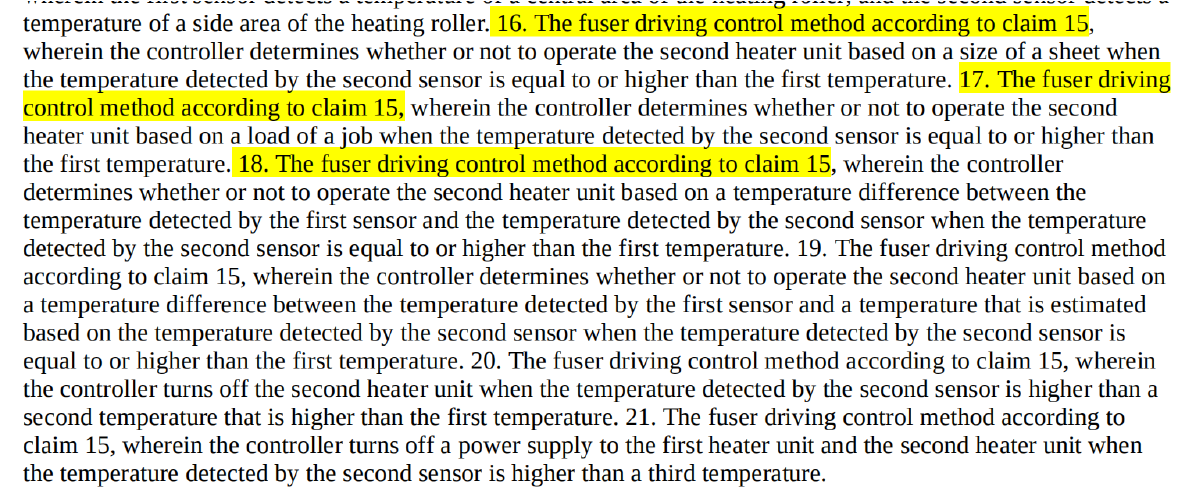}}
  \caption {A case for repetition error of patent generated by SFT.}
   \label{repetition-case}
\end{figure*}

patents generated using SFT exhibit significant repetition errors, whereas our AutoPatent framework produces more coherent content, thanks to the examiner agent. As shown in Figure~\ref{repetition-case}, we display an example of the repetition error in the patent generated by SFT. We can observe that the claims in the figure are meaningless, while in the corresponding real patent, there are only ten claims.

\subsection{Comprehensiveness Case}
\label{D2}
\begin{figure*}[htpb]
\centering
  \fbox{\includegraphics[width=1\linewidth]{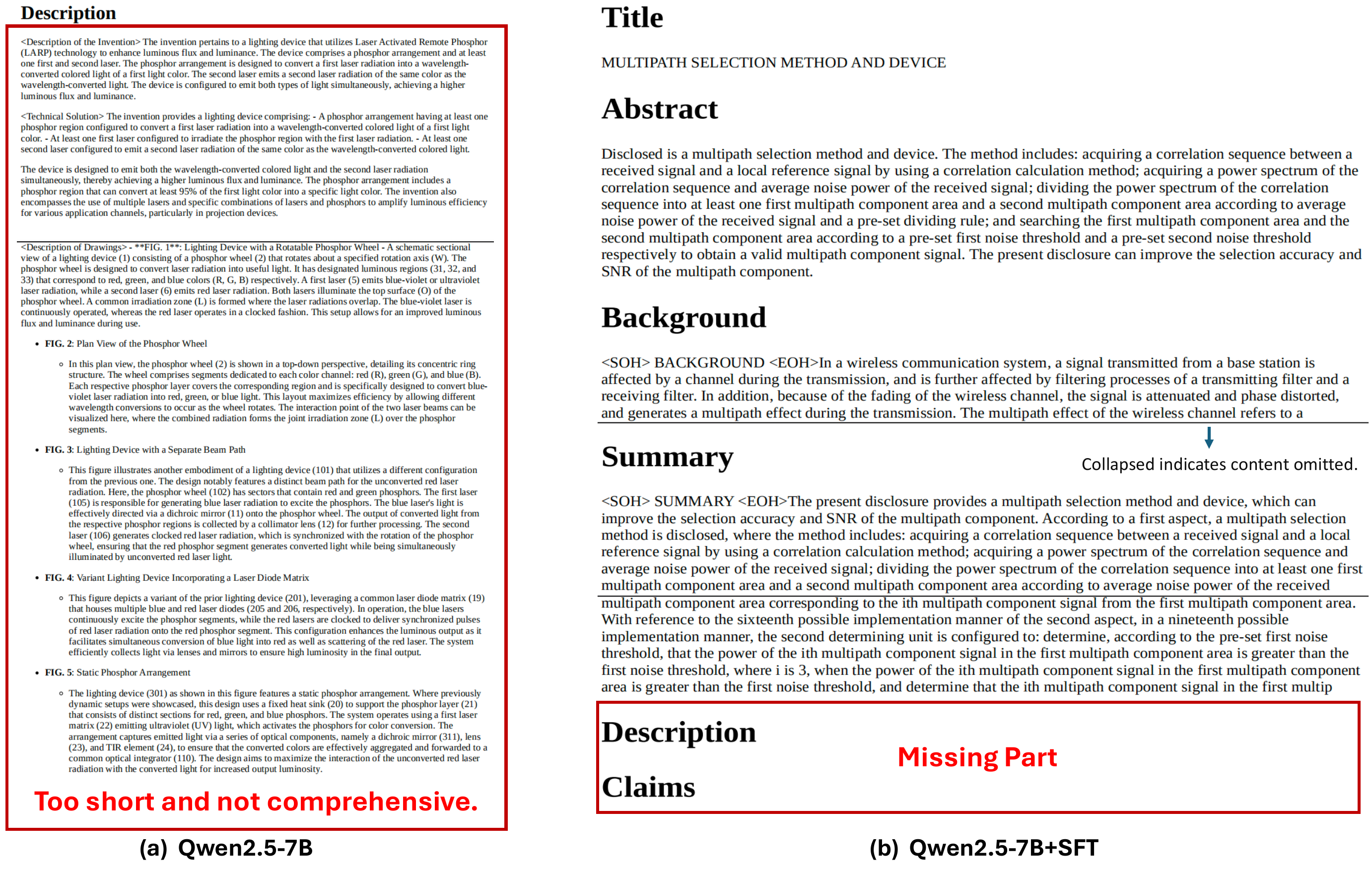}}
  \caption {A case for patent’s comprehensiveness generated using (a) zero-shot prompting and (b) Supervised Fine-Tuning.}
   \label{missingpart-case}
\end{figure*}

\begin{figure*}[htpb]
\centering
  \fbox{\includegraphics[width=1\linewidth]{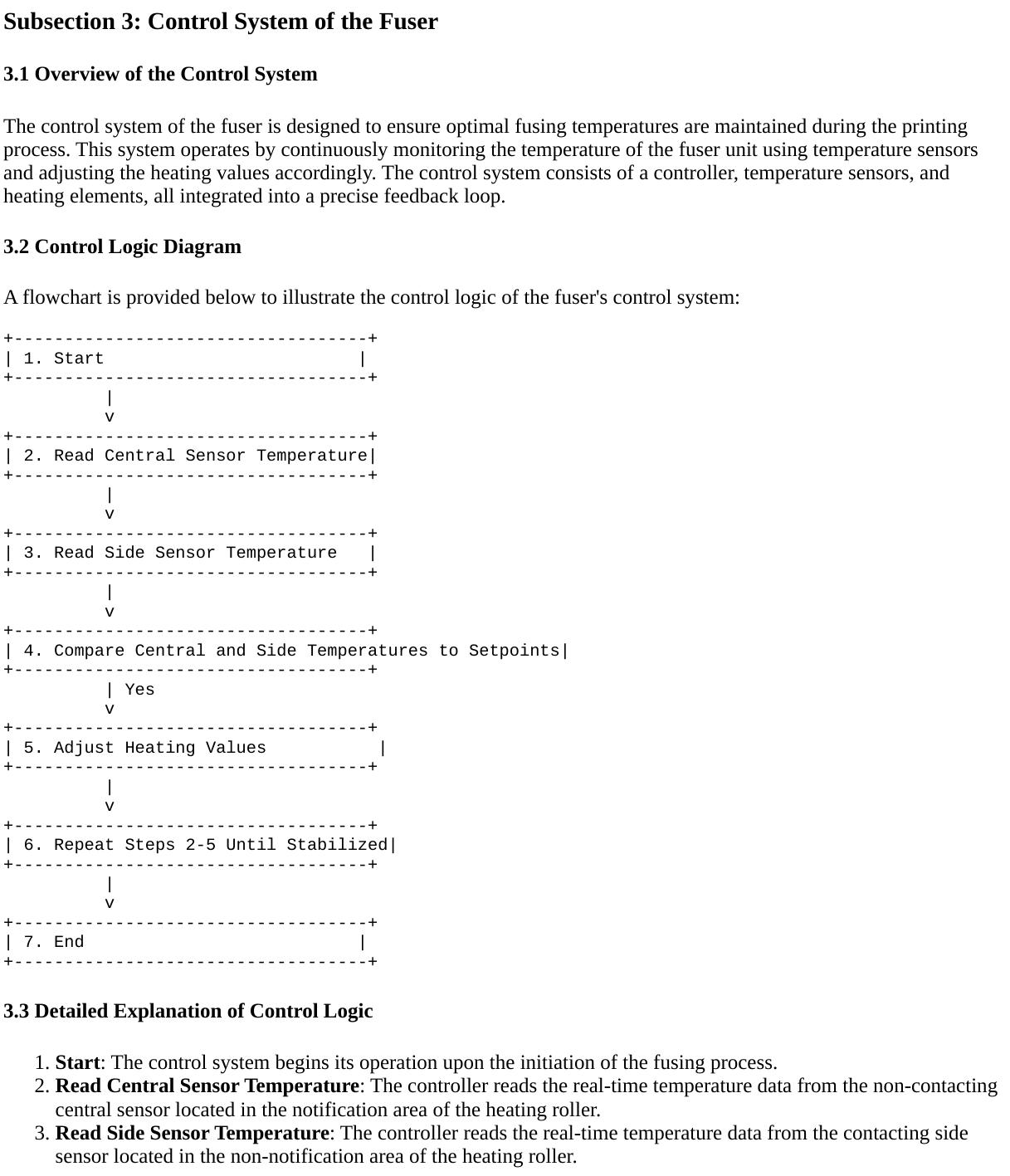}}
  \caption {A case for patent’s comprehensiveness generated using AutoPatent framework.}
   \label{flowchart-case}
\end{figure*}

patents generated using the AutoPatent framework are more comprehensive than those produced by other methods, which often fail to generate a complete patent. As shown in Figure~\ref{missingpart-case}, we display two examples where zero-shot prompting and SFT fail to generate some parts of the patent. We can observe that the zero-shot prompting method can generate all parts of the patent, but the content is shallow and lacks depth and comprehensiveness, simplifying complex ideas. And the patent generated using SFT even fails to generate the description and claims. As shown in Figure~\ref{flowchart-case}, we observe that patents generated using the AutoPatent framework can even include flowcharts, a capability attributed to the planning agent.

\end{document}